\documentclass[letterpaper]{article} 
\usepackage{amssymb}
\usepackage{amsmath}
\usepackage[preprint]{aaai2027}  
\usepackage[hyphens]{url}  
\usepackage{booktabs}
\usepackage{threeparttable}
\usepackage{graphicx} 
\usepackage{natbib}  
\usepackage{caption} 
\usepackage{xspace}
\usepackage{amsmath}
\usepackage{amssymb} 
\usepackage{algorithm}
\usepackage{algpseudocode}
\usepackage{booktabs}
\usepackage[most]{tcolorbox}
\usepackage[table]{xcolor}

\definecolor{bluex}{HTML}{2F65A7}
\newtcolorbox{promptbox}[1]{
  enhanced,
  breakable,
  boxrule=0.8pt,
  arc=4pt,
  fonttitle=\bfseries\small\rmfamily,
  fontupper=\small\rmfamily,
  before upper={\small\rmfamily\setlength{\parskip}{4pt}},
  coltitle=white,
  colbacktitle=bluex,
  colback=bluex!3,
  colframe=bluex!70,
  left=4pt,
  right=4pt,
  top=4pt,
  bottom=4pt,
  before skip=8pt,
  after skip=8pt,
  title={#1}
}
\newtcolorbox{contractbox}[1]{
  enhanced,
  breakable,
  boxrule=0.5pt,
  arc=2pt,
  fonttitle=\bfseries\small\rmfamily,
  fontupper=\small\rmfamily,
  before upper={\small\rmfamily\setlength{\parskip}{3pt}},
  coltitle=bluex,
  colbacktitle=bluex!8,
  colback=white,
  colframe=bluex!45,
  left=3pt,
  right=3pt,
  top=3pt,
  bottom=3pt,
  before skip=5pt,
  after skip=5pt,
  title={#1}
}
\newcommand{\promptterm}[1]{\textnormal{#1}}
\newcommand{\contractfield}[2]{\noindent\textbf{#1.}\ #2\par}

\title{SkillHEX: Improving Agent Skills via Hypothesis-Driven \\ Autonomous Exploration and Exploitation}
\author{
    Yuru Feng\textsuperscript{\rm 1,\rm 2},
    Yaoqi Chen\textsuperscript{\rm 1,\rm 3},
    Beidi Zhao\textsuperscript{\rm 1,\rm 4},
    Qianxi Zhang\textsuperscript{\rm 1},
    Xinjiang Wang\textsuperscript{\rm 1},\\
    Jianan Lu\textsuperscript{\rm 1},
    Zhirui Wang\textsuperscript{\rm 1},
    Shusen Xu\textsuperscript{\rm 1},
    Zengzhong Li\textsuperscript{\rm 1},
    Qi Chen\textsuperscript{\rm 1}
}
\affiliations{
    \textsuperscript{\rm 1}Microsoft, 
    \textsuperscript{\rm 2}University of California, San Diego,\\
    \textsuperscript{\rm 3}University of Science and Technology of China,
    \textsuperscript{\rm 4}University of British Columbia
}

\begin{document}

\maketitle

\begin{abstract}
Although agent skills equip LLMs with reusable procedural knowledge, manual maintenance suffers from high costs, unscalability, and misalignment. Real-world deployments thus require autonomous, on-demand skill evolution at test time, constrained by limited interaction budgets and a lack of training or validation sets. This setting introduces a severe \emph{sparse reward} challenge, where outcomes conflate multiple latent failure causes. Under such ambiguity, existing methods that greedily refine a single incumbent skill are particularly vulnerable to an \emph{exploitation trap}, allowing early misdiagnoses to exhaust limited trials along unproductive trajectories. To address this, we introduce SkillHEX, a closed-loop framework coupling hypothesis-driven self-verification with evidence-guided tree search. SkillHEX translates falsifiable failure hypotheses into executable tests, producing diagnostic evidence as dense reward without additional environment attempts. This evidence guides a search over persistent skill-revision branches, dynamically balancing the exploitation of supported edits with the exploration of plausible alternatives. Evaluated on $87$ tasks from SkillsBench, SkillHEX outperforms existing self-evolving methods and achieves an average pass rate of $55.9\%$ and $57.9\%$ using GPT-5.3-Codex and Claude Opus 4.7 under a five-iteration budget, respectively.

\end{abstract}

\section{Introduction}
Truly autonomous agents should do more than execute a fixed procedure: when an execution fails, a capable agent should identify what may have gone wrong, revise its strategy, and systematically apply the resulting knowledge to subsequent executions. Agent skills provide a practical interface for this capability because they package reusable procedures, scripts, and reference materials as modular, isolated, and editable artifacts that can be selectively loaded at inference time~\cite{min2025quco,zhang2024agent,zou2025llm,huang2026rethinking,zou2026when,anthropic2025agentskills,xu2026agent}. However, manually constructing and maintaining an expanding collection of skills is costly, difficult to scale, and heavily bottlenecked by the inherent human--machine reasoning gap \cite{coevoskills,skillsbench}. These limitations motivate \emph{automated skill evolution}: enabling an agent to revise its reusable procedural knowledge using evidence from its own executions~\cite{agentskillevaluationevolution}. 

Existing approaches to self-evolving skills generally follow one of two paradigms: 1) \emph{text-space optimization}, which treats skill instructions as an optimizable parameter space to compute textual language
gradients~\cite{TextGrad,gepa,skillopt, skillevolver}; or 2) \emph{experience-based approaches}, which distill procedural heuristics from execution traces or cross-task trajectories~\cite{trace2skill,skillrl,evoskill,skillos,skillforge,skillfoundry}. By repeatedly testing candidate skill revisions on \emph{training or validation sets}, both paradigms obtain dense feedback to steer the learning process, allowing them to systematically accept superior updates and reject regressions over evolution cycles.

However, unlike traditional paradigms reliant on pre-collected training and validation sets, real-world agent deployments face a strictly on-demand scenario. Lacking prior datasets, agents are afforded only a limited number of exploration trials before a skill must be deployed. This test-time adaptation setting introduces a fundamental \emph{sparse reward} challenge: forced to depend on sparse execution outcomes instead of dense validation feedback, agents face significantly exacerbated difficulty in selecting appropriate candidate skill revisions. 

\begin{figure}[t]
\centering
\includegraphics[width=0.85\columnwidth]{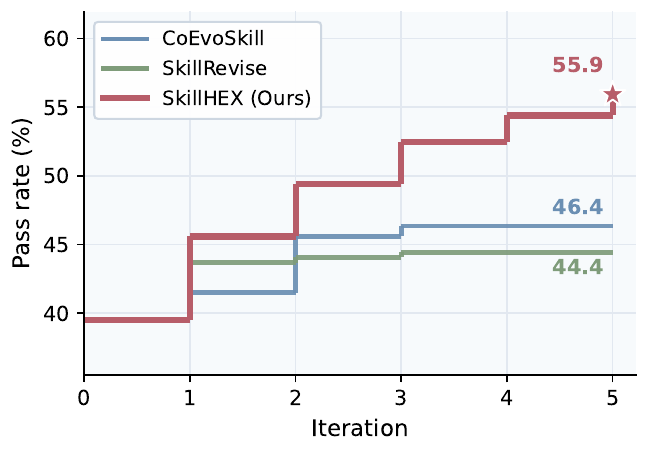}
\caption{\textbf{Evolution trajectory comparison on SkillsBench using GPT-5.3-Codex.} Baseline self-refinement methods (CoEvoSkills and SkillRevise) quickly plateau due to early commit traps, whereas SkillHEX achieves continuous performance gains across limited trial budgets.}
\label{fig:evolution-trajectory}
\end{figure}

This sparse reward challenge is particularly severe because environmental feedback often collapses many latent failure causes into the same ambiguous observation. While recent self-supervised refinement frameworks~\cite{coevoskills, skillrevise} attempt to adapt by diagnosing skill defects from execution evidence and applying targeted edits, their underlying revision selection mechanisms remain fundamentally unchanged from previous approaches. Specifically, these methods remain inherently exploitation-driven at the skill-revision level, as each iteration commits to the step-level preferred edit of current skills. Under such ambiguity, this greedy in-place refinement is highly vulnerable: it commits each patch to a single trajectory, allowing an early misdiagnosis to exhaust the limited trials along an unproductive direction (Figure~\ref{fig:evolution-trajectory}). 

We argue that effectively evolving skills at test time requires delicately balancing \emph{exploitation} and \emph{exploration}, which introduces two specific challenges. First,  \textbf{the exploitation trap}: due to sparse rewards and biased self-supervision signals, greedily exploiting a noise-corrupted diagnosis may trap the agent in local optima by converging on flawed patches. Second, \textbf{the exploration constraint}: the vast space of possible skill revisions cannot be exhaustively searched given the high cost of environment interactions. Therefore, an efficient search mechanism must actively gather reliable evidence and preserve alternative revision paths to prevent premature commitment, while operating under a bounded cost and interaction budget.

Motivated by this observation, we introduce SkillHEX, a closed-loop framework that couples hypothesis-driven self-verification with evidence-guided tree search. To overcome the \emph{exploitation trap} caused by sparse feedback, SkillHEX introduces a \textbf{self-verifier}. It turns explicit, falsifiable failure-cause hypotheses into validated executable tests and reruns new tests on agents' answers, producing dense evidence without additional executor attempts. To address the \emph{exploration constraint}, SkillHEX employs an \textbf{evidence-guided tree search}. The search process is directly guided by the evidence produced by the self-verifier, dynamically balancing the exploitation of evidence-supported branches with the exploration of plausible but underexplored alternatives. Consequently, unlike existing greedy approaches that plateau after early iterations, SkillHEX achieves sustained performance improvement across limited trials (Figure~\ref{fig:evolution-trajectory}).

In summary, our contributions are threefold. First, we introduce hypothesis-driven self-verification, which converts explicit, falsifiable failure-cause hypotheses into validated executable tests and aggregates their outcomes into dense, dynamically updated evidence for skill revision. Second, we develop an evidence-guided tree search over persistent skill-patch branches, allowing the budget allocator to revisit plausible alternatives instead of forcing every later revision to inherit an early diagnosis. Finally, experiments on $87$ SkillsBench tasks across two backbones show gains of up to $9.5$ percentage points over the strongest self-evolution baseline.

\section{Related Work}
\paragraph{Agent Skills and Automated Skill Evolution.}
Agent skills package instructions, scripts, and reference materials as reusable artifacts that agents load at inference time~\cite{anthropic2025agentskills,xu2026agent}. Related text-space optimizers derive instruction updates from repeated language-model feedback~\cite{TextGrad,gepa,skillopt}, while experience-based methods distill procedures from execution trajectories~\cite{trace2skill,skillrl,evoskill}. Methods closer to our setting revise skills using task-specific executions: SkillEvolver refines domain skills by evaluating fresh-agent deployments across a provided training split~\cite{skillevolver}, CoEvoSkills co-evolves skill generation with surrogate verification~\cite{coevoskills}, and SkillRevise sequentially refines an incumbent skill from execution traces~\cite{skillrevise}. However, these existing approaches fail to account for the critical need to balance exploration and exploitation in new scenarios. SkillHEX simultaneously address the exploitation trap and the exploration constraint through hypothesis-targeted evidence and an evidence-guided tree search.

\paragraph{Self-Generated Verification and Search.}
Self-refinement and LLM-as-a-judge methods provide inexpensive test-time feedback, but model-generated judgments can amplify self-bias and exhibit systematic evaluation biases~\cite{selfbias,llmjudge}. Tree-search methods such as LATS and rStar preserve alternatives while balancing exploration and exploitation over reasoning or action states~\cite{lats,rstar}; APEX applies a related idea to policy-level exploration in self-evolving agents~\cite{apex}, and iterative code repair exposes a similar allocation trade-off among program candidates~\cite{coderepair}. SkillHEX combines these directions by expressing suspected skill defects as falsifiable hypotheses, validating generated executable tests, replaying them on cached outputs, and using the resulting evidence in a PUCT-style search~\cite{multiarmed,puct}. Its search state is a persistent tree of reusable skill revisions rather than a transient reasoning trace or candidate program.

\section{Methodology}
\paragraph{Task Definition.} 
We consider a task $\mathcal{T} = (I, D, \mathcal{E})$, defined by an instruction $I$, public task data $D$, and an execution environment $\mathcal{E}$. The exact success criteria $\xi$ are hidden from the agent and implemented by a ground-truth verifier $V_{\xi}$, so the interaction induced by $\mathcal{E}$ is partially observable. At each step $t$ in one attempt, the agent issues an action $a_t$ (e.g., commands or file edits) and receives a partial observation $o_t$ (e.g., execution results). Consequently, the agent must act based on the historical trajectory of observations and actions, denoted as $h_t = (o_0, a_0, \ldots, a_{t-1}, o_t)$. Upon completion, rather than dense step-wise rewards, the agent receives a binary terminal reward from the verifier $V_{\xi}$, which evaluates the final outputs against the hidden criteria.

Given that skills $S$ condition the agent's behavior, our method focuses directly on its evolution. During execution, the LLM-based agent policy $\pi_{\mathrm{agent}}$ generates actions dynamically:

\begin{equation}
a_t \sim \pi_{\mathrm{agent}}\!\left(\cdot \mid \mathcal{T}, h_t, S\right)
\end{equation}

This policy induces an execution trajectory $\tau \sim P_{\mathrm{exec}}(\cdot \mid I, D, \mathcal{E}, S; \pi_{\mathrm{agent}})$, resulting in terminal output files $Y(\tau)$ generated by the agent. At evolution round $k$, evaluating the current skill version $S^{(k)}$ produces a trajectory $\tau^{(k)}$ and terminal outputs $Y^{(k)} = Y(\tau^{(k)})$. The ground-truth verifier assesses the final outputs to provide a reward: $R_k = V_{\xi}\!\left(Y^{(k)}\right) \in \{0,1\}$, where $1$ denotes task success and $0$ denotes failure. 

Beginning with an initial skill $S^{(0)}$, the evolution policy $\pi_{\mathrm{evo}}$ iteratively distills procedural knowledge from execution traces to update the skill versions. The proposal of the next candidate skill from the space of admissible skills $\mathcal{S}$ by the meta-policy can be formulated as:
\begin{equation}
    S^{(k+1)} \sim \pi_{\mathrm{evo}}\!\left(\cdot \mid \mathcal{T}, S^{(k)}, Y^{(k)}, R_k\right)
\end{equation}

To formalize the objective of this evolution process, we define the expected performance of any candidate skill $S \in \mathcal{S}$ on task $\mathcal{T}$ as its expected reward:
\begin{equation}
    G_{\mathcal{T}}(S) \triangleq \mathbb{E}_{\tau\sim P_{\mathrm{exec}}(\cdot \mid \mathcal{T}, S; \pi_{\mathrm{agent}})}\!\left[V_{\xi}(Y(\tau))\right]
\end{equation}
Ultimately, the overarching objective of the evolution policy is to discover an optimal skill $S^*$ that maximizes this expected success probability:
\begin{equation}
    S^*=\arg\max_{S\in\mathcal{S}} G_{\mathcal{T}}(S)
    \label{eq:optimization}
\end{equation}

\begin{figure*}[t]
\centering
\includegraphics[width=0.9\textwidth]{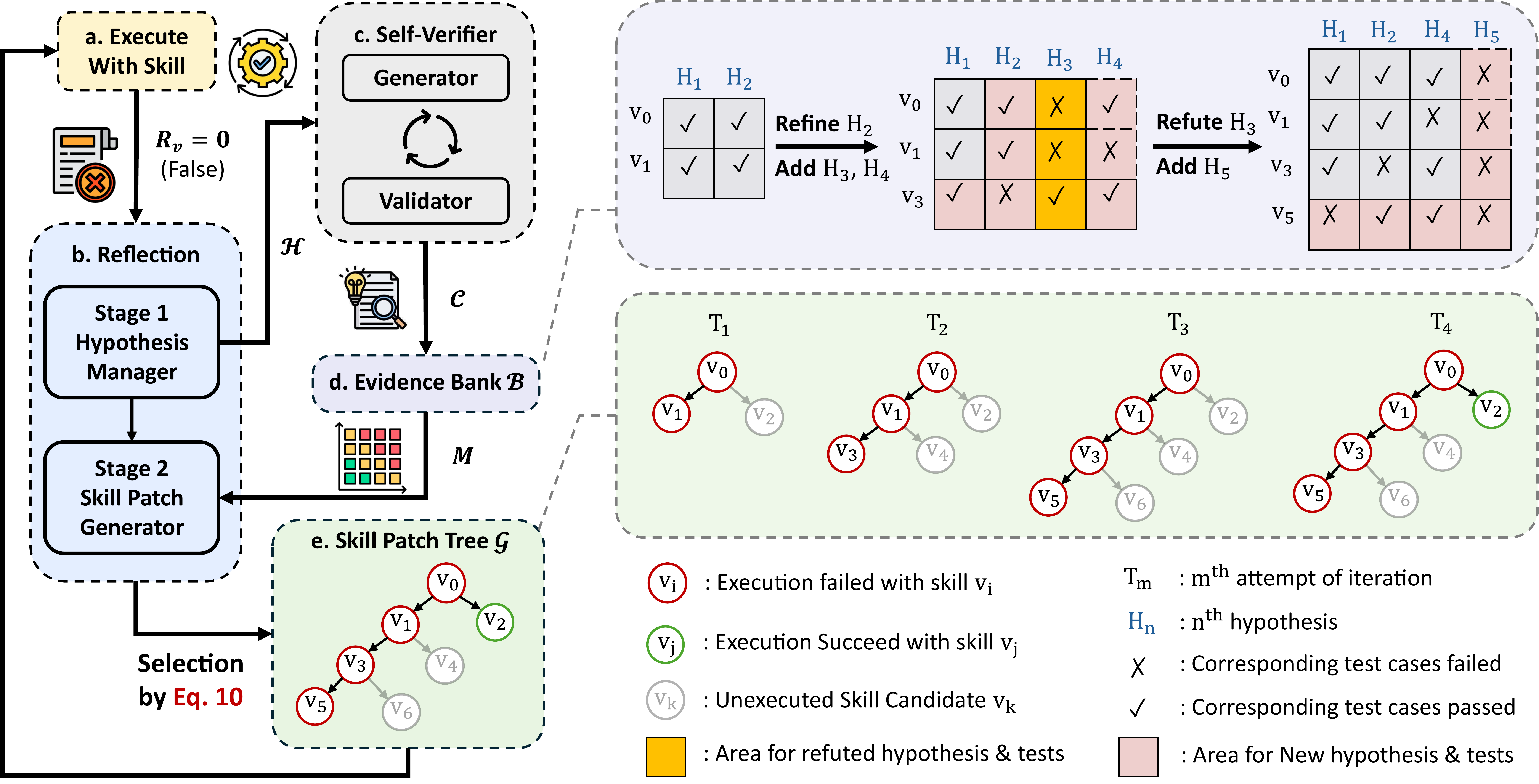} 
\caption{\textbf{Overview of SkillHEX.} (a) A selected skill version is executed on the task, where a failure outcome ($R_v = 0$) triggers the subsequent pipeline. (b) Reflection proceeds in two stages: it first formulates targeted, falsifiable failure hypotheses ($\mathcal{H}$) and iterates with the self-verifier until the evidence ($\mathcal{B}$) is judged sufficient or the evidence-round limit is reached; it then synthesizes candidate skill revisions. (c) For each active hypothesis $h \in \mathcal{H}$, the self-verifier's generator creates executable test cases ($\mathcal{C}$); the validator assesses their quality and returns the defective ones to the generator for another generation round. (d) The validated cases are run on cached outputs to update shared evidence matrix ($\mathcal{M}$) across hypotheses and evaluated skill versions. (e) Evidence-guided tree search ($\mathcal{G}$) preserves skill patches and selects the candidate for the next iteration.}

\label{fig:method}
\end{figure*}

\paragraph{SkillHEX Framework.}
\label{sec:overview}
Since directly optimizing $G_{\mathcal{T}}(S)$ (Equation~\ref{eq:optimization}) under sparse rewards is intractable, we introduce a hypothesis-driven self-verifier to address this challenge. By leveraging the reasoning capabilities of LLMs, this module generates dense, self-supervised diagnostic signals that effectively act as "semantic gradients." Coupled with this, we propose a tree-based search mechanism that uses these gradients to guide skill revisions. Together, this dual-component architecture efficiently balances exploration and exploitation across the vast discrete search space.

At evolution round $k$, the framework maintains four shared states to bridge its two core components. The skill-patch tree $\mathcal{G}_k$ stores each skill version $S_v$ as a node $v\in V(\mathcal{G}_k)$, caching agent output files $Y_v$ and corresponding rewards $R_v$ for evaluated nodes. The hypothesis store $\mathcal{H}_k$ tracks falsifiable failure causes, which are explicitly linked to executable tests in the validated test bank $\mathcal{C}_k$. The evidence bank $\mathcal{B}_k$ aggregates the outcomes derived from executing these tests, dynamically curating this accumulated evidence across evolution rounds. Operationally, the self-verifier continuously updates $\mathcal{H}_k$, $\mathcal{C}_k$, and $\mathcal{B}_k$, supplying the accumulated evidence that the search mechanism leverages to efficiently expand and select nodes within $\mathcal{G}_k$, as shown in Figure~\ref{fig:method}.

\subsection{Hypothesis-Driven Self-Verification}
\label{sec:selfverify}

To overcome the \emph{exploitation trap} caused by sparse rewards, SkillHEX introduces \textbf{Hypothesis-Driven Self-Verification}. This module operates through three main components: (1) actively managing hypotheses to track potential failure causes, (2) generating and validating tests based on these hypotheses, and (3) maintaining a dynamic evidence bank that provides a continuously evolving, dense diagnostic signal to guide skill revisions. These components are detailed in the following paragraphs.

\paragraph{Hypothesis Management.}
A task-level failure yields a strictly sparse binary signal, inherently collapsing a multitude of latent failure causes into the exact same observation. This severe partial observability confounds credit assignment, making it impossible to isolate which specific instruction, procedure, or constraint within a skill is responsible. To address this, we represent candidate failure causes as a dynamically maintained set of hypotheses $\mathcal{H}_k=\{h_i\}_{i=1}^{m_k}$. Each hypothesis is parameterized as a tuple $h_i=(d_i, q_i, \sigma_i, \mathcal{C}_i)$, where $d_i$ is a falsifiable description of the suspected defect, $q_i$ defines the observable behavior required to support or refute it, $\sigma_i$ denotes its lifecycle state (e.g., active, refuted), and $\mathcal{C}_i$ is the subset of tests linked to it. Guided by the current skill version, its execution trace, and historical test outcomes, the reflection agent executes three hypothesis-management operations: \textsc{Add} (proposing a new hypothesis), \textsc{Refine} (updating an existing hypothesis), and \textsc{Refute} (discarding an invalid cause), driving the evidence bank evolution process illustrated in Figure~\ref{fig:method}.

\paragraph{Test Generation and Validation.}
When existing evidence is insufficient to confidently propose a patch, the reflection mechanism selects a subset of active hypotheses $\mathcal{A}_k \subseteq \mathcal{H}_k$. For each $h_i \in \mathcal{A}_k$, the self-verifier's generator synthesizes executable tests targeting its predicted behavior $q_i$. Each generated test is explicitly tagged with the hypothesis it probes and assigned an assertion strength based on its test category. These tests subsequently pass through a rule-based validator to ensure syntactic correctness and executability. This generation-validation process operates iteratively, allowing the reflection agent and the self-verifier to interact for several rounds until sufficient diagnostic evidence is accumulated to inform skill revisions.

\paragraph{Dynamic evidence bank.}
To derive self-supervised guidance for the evolution process and actualize the utility of "semantic gradients", the framework maintains a dynamic evidence bank. This bank maintains a continuously evolving evidence matrix $M$, which serves as the diagnostic bridge connecting hypothesis-driven tests with executed revisions. For every evaluated node $v \in V^{\mathrm{eval}}$ and validated test $c_j \in \mathcal{C}$, the outcome is recorded as: 
\begin{equation}
M[v,j]=\phi(c_j, Y_v) \in \{0, 1\},
\end{equation}
where $\phi$ executes test $c_j$ against the terminal output files $Y_v$ cached by node $v$, with values $1$ and $0$ denoting pass and fail, respectively.

The evidence matrix actively co-evolves with the evolutionary search, enabling continuous cross-validation along both axes. Evaluating a newly proposed patch expands the frontier by appending a \emph{row}, executing every existing test against its novel output. Orthogonally, validating a new hypothesis-driven test appends a \emph{column} by retrospectively replaying it across all historically cached executions:
\begin{equation}
\begin{aligned}
M_{k+1}[v, :] &= \{\phi(c_j, Y_v) \mid c_j \in \mathcal{C}_k\}, \\
M_{k+1}[:, j] &= \{\phi(c_j, Y_v) \mid v \in V_k^{\mathrm{eval}}\}.
\end{aligned}
\end{equation}
This symmetric expansion is highly efficient: a single new test instantly refreshes the diagnostic landscape for all prior revisions, while every new revision is immediately subjected to the rigorous constraints of the entire historical tests.

Although self-supervised diagnostics address reward sparsity, they inherently suffer from the reasoning biases of the LLM, which may generate flawed hypotheses and tests. To maintain the fidelity of the derived semantic gradients, the evidence bank actively prunes misleading signals. When a hypothesis $h_i$ is designated as $\mathrm{refuted}$, its associated tests become obsolete, triggering the removal of their corresponding columns from the matrix:
\begin{equation}
 \mathcal{D}_i = \mathcal{C}_i \setminus \bigcup_{h_\ell:\, \sigma_\ell \neq \mathrm{refuted}} \mathcal{C}_\ell,\; M \leftarrow M[:,\, \mathcal{C} \setminus \mathcal{D}_i].
 \end{equation}
Through this combination of dynamic expansion and targeted pruning, the evidence bank provides diagnostic signals that are denser and more reliable than those used in previous methods, thereby offering superior guidance for the evolution policy.

\subsection{Evidence-Guided Tree Search}
\label{sec:search}
To address the \emph{exploration constraint}, we formulate the evolution policy $\pi_{\mathrm{evo}}$ as a persistent patch tree search over $\mathcal{G}_k$. Unlike approaches that greedily revise in-place, SkillHEX balances the exploitation of evidence-supported branches with the exploration of plausible alternatives. The following paragraphs detail the mechanisms of this search process.

\paragraph{Semantic Expansion via Reflection Prior.}
The root of $\mathcal{G}_k$ initializes at $S^{(0)}$. When an evaluated node $v$ (representing skill version $S_v$) is selected for expansion, we leverage the reflection agent to act as an evidence-conditioned transition dynamics. Drawing upon the active hypotheses $\mathcal{H}_k$ and execution evidence $\mathcal{B}_k$, the agent proposes up to $K$ candidate patches. 
Rather than relying on poorly calibrated absolute confidence scores from the language model, we ask the reflection agent to provide an ordinal ranking $\rho_u \in \{1, \dots, K\}$ for each proposed child $u$. We map this ordinal consensus into a normalized policy prior:
\begin{equation}
P(v,u) = \frac{\exp[-\beta(\rho_u-1)]}{\sum_{w\in\mathrm{Ch}(v)}\exp[-\beta(\rho_w-1)]},
\label{eq:tree-prior}
\end{equation}
where $\mathrm{Ch}(v)$ denotes the children of $v$, and $\beta$ is a temperature hyperparameter controlling the reliance on the reflection ranking. Crucially, all valid children are preserved in $\mathcal{G}_k$. This prior biases the search toward the most promising hypotheses without irreversibly pruning lower-ranked, yet potentially viable, alternatives.

\paragraph{Hierarchical Evaluation and Max-Backup.}
To evaluate a newly proposed node $v$ in the search tree, we recognize that not all tests are equally informative. For instance, if a skill revision introduces a basic syntax or formatting error, evaluating its performance on complex functional metrics becomes meaningless. To address this, we implement a strict hierarchical gating mechanism by categorizing tests into hard constraints (e.g., syntax validity, deterministic output formats) and semantic metrics, which jointly determine the node's evaluation score $s(v)$. By conditioning the semantic metrics on the perfect satisfaction of all hard constraints, this mechanism naturally steers the search to prioritize rectifying fundamental errors for the most direct optimization gains.

Since our primary objective is to extract the most effective skill variant discovered during the search, we employ a \textit{max-backup} operator. The backed-up value $Q(v)$ of an internal node $v$ reflects the strongest empirical evidence discovered anywhere in its subtree:
\begin{equation}
Q(v) = \max\!\left(s(v), \max_{u\in\mathrm{Ch}_{\mathrm{eval}}(v)}Q(u)\right),
\label{eq:max-backup}
\end{equation}
where $\mathrm{Ch}_{\mathrm{eval}}(v)$ is the set of its evaluated children. This formulation ensures that unsuccessful exploratory branches do not penalize the value of a strong parent node.

\paragraph{Evidence-Driven Selection.}
To proactively escape the exploitation trap and balance the trade-off between exploiting highly-valued nodes and exploring alternative candidates, we adopt a variant of the PUCT selection rule~\cite{multiarmed,puct}, replacing its conventional mean edge value with our max-backup value. We dynamically normalize the max-backup values into $\bar{Q}(u) \in [0,1]$ using global min-max bounds. Starting from the root, the search recursively selects the child $u^*$ that maximizes the upper confidence bound:
\begin{equation}
u^* = \arg\max_{u \in \text{Ch}(v)} \left[ \bar{Q}(u) + c_{\text{puct}} P(v, u) \frac{\sqrt{\max\{1, N(v)\}}}{1 + N(v, u)} \right],
\end{equation}
where $N(v,u)$ is the number of times edge $(v,u)$ has been selected during tree traversal, $N(v)=\sum_{w\in\mathrm{Ch}(v)}N(v,w)$ is the total number of outgoing-edge traversals from $v$. The $\max\{1,N(v)\}$ convention keeps the exploration bonus nonzero before any child edge has been traversed. The traversal stops at an unexpanded or unevaluated node, triggering either a new reflection expansion or an executor evaluation. By explicitly coupling semantic priors $P(v,u)$ with empirical backups $\bar{Q}(u)$, the persistent tree gracefully shifts to alternative revision paths whenever new evidence diminishes the estimated value of the current trajectory, thereby effectively balancing exploitation and exploration.

\begin{table*}[t]
\centering
\setlength{\tabcolsep}{1mm}
\begin{threeparttable}
\begin{tabular}{lcccc>{\columncolor{blue!10}}cc}
\toprule
Domain & No Skill & Skill Creator & CoEvoSkills & SkillRevise
& SkillHEX (Ours) & \emph{Human} \\
\midrule
\multicolumn{7}{c}{\textbf{\textit{GPT-5.3-Codex}}} \\
\midrule
Software Engineering
& $14.6_{\pm 3.6}$ & $33.3_{\pm 3.6}$ & $31.2_{\pm 6.2}$
& $31.2_{\pm 6.2}$ & $\mathbf{52.1}_{\pm 3.6}$ & $47.9_{\pm 7.2}$ \\
Cybersecurity
& $47.6_{\pm 8.2}$ & $42.9_{\pm 14.3}$ & $61.9_{\pm 8.2}$
& $61.9_{\pm 8.2}$ & $\mathbf{71.4}_{\pm 0.0}$
& $\mathbf{71.4}_{\pm 0.0}$ \\
Natural Science
& $38.1_{\pm 21.8}$ & $54.8_{\pm 4.1}$ & $59.5_{\pm 4.1}$
& $54.8_{\pm 4.1}$ & $61.9_{\pm 4.1}$ & $\mathbf{76.2}_{\pm 10.9}$ \\
Finance \& Economics
& $29.6_{\pm 6.4}$ & $25.9_{\pm 6.4}$ & $25.9_{\pm 6.4}$
& $33.3_{\pm 0.0}$ & $29.6_{\pm 6.4}$ & $\mathbf{37.0}_{\pm 6.4}$ \\
Office \& White Collar
& $47.6_{\pm 4.1}$ & $40.5_{\pm 4.1}$ & $52.4_{\pm 4.1}$
& $42.9_{\pm 0.0}$ & $\mathbf{61.9}_{\pm 8.2}$ & $42.9_{\pm 0.0}$ \\
Media \& Content Production
& $40.0_{\pm 0.0}$ & $26.7_{\pm 11.5}$ & $53.3_{\pm 30.6}$
& $46.7_{\pm 11.5}$ & $53.3_{\pm 11.5}$ & $\mathbf{60.0}_{\pm 20.0}$ \\
Industrial \& Physical Systems
& $31.0_{\pm 4.1}$ & $31.0_{\pm 4.1}$ & $40.5_{\pm 8.2}$
& $38.1_{\pm 14.9}$ & $42.9_{\pm 7.1}$ & $\mathbf{45.2}_{\pm 8.2}$ \\
Mathematics \& OR
& $29.2_{\pm 7.2}$ & $58.3_{\pm 7.2}$ & $58.3_{\pm 7.2}$
& $62.5_{\pm 0.0}$ & $\mathbf{83.3}_{\pm 7.2}$ & $50.0_{\pm 12.5}$ \\
\midrule
\rowcolor{gray!15} \textbf{Overall}
& $33.3_{\pm 4.1}$ & $39.5_{\pm 1.8}$ & $46.4_{\pm 0.7}$
& $44.4_{\pm 3.7}$ & $\mathbf{55.9}_{\pm 1.8}$ & $52.9_{\pm 5.0}$ \\
$\Delta$ vs.\ No Skill (pp)
& -- & $+6.2$ & $+13.1$ & $+11.1$ & $\mathbf{+22.6}$ & $+19.6$ \\
\midrule
\multicolumn{7}{c}{\textbf{\textit{Claude Opus 4.7}}} \\
\midrule
Software Engineering
& $31.3_{\pm 6.3}$ & $31.3_{\pm 6.3}$ & $43.8_{\pm 6.3}$ & $43.8_{\pm 6.3}$
& $\mathbf{47.9}_{\pm 3.0}$ & $\mathbf{47.9}_{\pm 9.5}$ \\
Cybersecurity
& $42.9_{\pm 14.3}$ & $47.6_{\pm 8.3}$ & $47.6_{\pm 8.3}$ & $47.6_{\pm 8.3}$
& $\mathbf{66.7}_{\pm 6.7}$ & $61.9_{\pm 8.2}$ \\
Natural Science
& $38.1_{\pm 8.2}$ & $54.8_{\pm 8.3}$ & $69.1_{\pm 4.1}$ & $59.5_{\pm 4.1}$
& $66.7_{\pm 3.4}$ & $\mathbf{81.0}_{\pm 4.1}$ \\
Finance \& Economics
& $25.9_{\pm 6.4}$ & $18.5_{\pm 6.4}$ & $22.2_{\pm 0.0}$ & $22.2_{\pm 0.0}$
& $40.7_{\pm 5.2}$ & $\mathbf{51.9}_{\pm 6.4}$ \\
Office \& White Collar
& $40.5_{\pm 4.1}$ & $45.2_{\pm 10.9}$ & $54.8_{\pm 8.3}$ & $57.1_{\pm 7.1}$
& $\mathbf{71.4}_{\pm 5.8}$ & $50.0_{\pm 7.1}$ \\
Media \& Content Production
& $20.0_{\pm 20.0}$ & $26.7_{\pm 11.6}$ & $40.0_{\pm 0.0}$ & $26.7_{\pm 11.6}$
& $40.0_{\pm 0.0}$ & $\mathbf{66.7}_{\pm 11.5}$ \\
Industrial \& Physical Systems
& $23.8_{\pm 10.9}$ & $30.9_{\pm 4.1}$ & $42.9_{\pm 12.4}$ & $33.3_{\pm 4.1}$
& $\mathbf{47.6}_{\pm 6.7}$ & $40.5_{\pm 10.9}$ \\
Mathematics \& OR
& $50.0_{\pm 0.0}$ & $54.2_{\pm 7.2}$ & $66.7_{\pm 7.2}$ & $54.2_{\pm 7.2}$
& $\mathbf{79.2}_{\pm 5.9}$ & $41.7_{\pm 7.2}$ \\
\midrule
\rowcolor{gray!15} \textbf{Overall}
& $34.1_{\pm 3.3}$ & $39.1_{\pm 4.1}$ & $49.4_{\pm 5.0}$ & $44.8_{\pm 2.3}$ & $\mathbf{57.9}_{\pm 2.4}$ & $54.4_{\pm 1.3}$ \\
 $\Delta$ vs.\ No Skill (pp)
& -- & $+5.0$  & $+15.3$  & $+10.7$ & $\mathbf{+23.8}$& $+20.3$ \\
\bottomrule
\end{tabular}
\end{threeparttable}
\caption{\textbf{Main results on all 87 SkillsBench tasks.} All results are averaged over three seeds, with standard deviations shown as subscripts. Evolution methods start from the same Skill Creator initialization and are evaluated over five iterations. Bold denotes the highest mean in each row.}
\label{tab:main}
\end{table*}

\section{Experiments}

\subsection{Experimental Setup}

\paragraph{Benchmark.}
We evaluate SkillHEX on SkillsBench~\cite{skillsbench}, which comprises $87$ tasks across eight domains: Software Engineering, Cybersecurity, Natural Science, Finance \& Economics, Office \& White Collar, Media \& Content Production, Industrial \& Physical Systems, and Mathematics \& OR. Each task runs in an isolated Docker container that pairs with deterministic verifiers. While conventional agent benchmarks~\cite{agentbench,webarena,osworld,appworld,mlebench,bigcodebench} typically evaluate the overall pass rate of a fixed, static model--harness stack, SkillsBench isolates the \emph{installed skill} as the primary experimental variable. By holding the base model, harness, task container, and verifier strictly constant, SkillsBench allows us to precisely measure the incremental pass-rate improvements driven solely by autonomous skill evolution. This controlled evaluation paradigm naturally aligns with our on-demand test-time adaptation setting, where each executor attempt starts from a fresh instance and the only varying factor is the dynamically generated skill.

\paragraph{Baselines and Initialization.}
We compare SkillHEX against two categories of baselines: static and self-evolving methods. \emph{No Skill} runs the executor without an installed skill, \emph{Skill Creator}~\cite{skillcreator} uses the unrevised LLM-authored skill, and \emph{Human} uses human-curated skills provided by SkillsBench. We additionally compare with two recent self-evolution methods: CoEvoSkills~\cite{coevoskills} (since its source code is not publicly available, we implement and evaluate its framework based on the details in the paper), which co-evolves skills with surrogate verification, and SkillRevise~\cite{skillrevise}, which performs trace-conditioned sequential revision. We initialize SkillHEX and all self-evolution methods with the same LLM-generated skills produced by \emph{Skill Creator}, controlling the same starting point for evolution. 

\paragraph{Model.}
We evaluate all methods with two LLM backbones from different model families, GPT-5.3-Codex~\cite{gpt53codex} and Claude Opus 4.7~\cite{claudeopus47}. All LLM roles---executor, self-verifier, and reflection---use medium as default reasoning effort. To faithfully measure the impact of different skills, we designed a custom, lightweight harness for the executor agent that isolates task success from framework-level interventions.

\paragraph{Metrics.}
Following the evaluation protocol of SkillsBench ~\cite{skillsbench}, our primary metric is the task-macro pass rate. Each experiment is independently repeated over 3 runs, and we report both the mean pass rate and the standard deviation (std) across these runs. For evolutionary methods, results are evaluated and reported as $\text{pass@5}$ (with a maximum budget of five iterations). Additionally, we report the percentage point ($\text{pp}$) gains relative to the \emph{No Skill} baseline to quantify relative improvements.

\subsection{Main Results}

\paragraph{Overall Performance.}
As summarized in Table~\ref{tab:main}, SkillHEX outperforms baselines across both backbone models. Specifically, SkillHEX achieves the highest overall pass rates of $55.9\%$ (GPT-5.3-Codex) and $57.9\%$ (Claude Opus 4.7). Compared to the strongest baseline CoEvoSkills, our approach delivers notable gains of $9.5$ pp and $8.5$ pp, respectively. Furthermore, SkillHEX substantially outperforms the \emph{No Skill} baseline by $22.6$ pp and $23.8$ pp, even surpassing human-level performance. Crucially, this superior performance validates the core design of SkillHEX: by leveraging \emph{Hypothesis-Driven Self-Verification} and \emph{Evidence-Guided Tree Search}, our approach effectively mitigates both the exploitation trap caused by sparse rewards and the exploration constraint under bounded budgets. These results demonstrate that the effectiveness of our evolutionary approach persists regardless of the backbone models.

\paragraph{Gains span diverse task domains.}
SkillHEX consistently outperforms all baselines across every evaluated domain on SkillsBench. Compared to human-curated skills, our method matches or significantly exceeds their performance in several domains including  Software Engineering, Cybersecurity, Office \& White Collar and Mathematics \& OR. The clearest gains over human-curated skills occur in Office \& White
Collar, where SkillHEX improves by $19.0$ and $21.4$ pp, and Mathematics \& OR, where the gains reach $33.3$ and $37.5$ pp under the two backbones. These domains are characterized by highly interdependent workflows that demand precise state tracking, compounding constraints, and orchestration of multi-step procedures. For example, in the Mathematics \& OR domain, tasks typically require navigating intricate feasibility constraints, objective functions, and solver choices. SkillHEX allows distinct skill patches to preserve revisions that optimize specific objectives or constraints. Concurrently, the shared evidence bank enables the reflection agent to integrate multiple pieces of diagnostic evidence into a unified formulation. Empirical results indicate that this integrated mechanism equips the agent with the capability to escape local optima and achieve success. 

Conversely, SkillHEX still falls short of human-curated skills in certain knowledge-intensive domains, such as Natural Science and Finance \& Economics, across both backbone models. We hypothesize that this performance gap stems from a knowledge-acquisition bottleneck rather than a limitation of the search mechanism itself. These domains heavily rely on specialized domain knowledge and established priors that are difficult to infer purely from basic task specifications or a handful of failed executions~\cite{skillsbench}. 


\subsection{Ablation Studies}
\begin{table}[t]
\centering
\setlength{\tabcolsep}{3pt}
\begin{tabular}{lcc}
\toprule
Method & Pass Rate (\%) & Drop (pp) \\
\midrule
SkillHEX (full)   & $\mathbf{55.9}_{\pm 1.8} $ & 0.0 \\
\quad w/o self-verifier & $44.8_{\pm 2.0}$ & $\downarrow 11.1$ \\
\quad w/o skill patch tree & $49.1_{\pm 1.3}$ & $\downarrow 6.8$ \\
\bottomrule
\end{tabular}
\caption{\textbf{Ablation results on SkillsBench using GPT-5.3-Codex over five iterations.} The method without the skill patch tree corresponds to in-place refinement. Impact is the
absolute drop from no ablation.}
\label{tab:ablation}
\end{table}
To quantify the contribution of each component in SkillHEX, we compare exactly three methods: no ablation, w/o self-verifier, w/o skill patch tree (in-place refinement). 
Table~\ref{tab:ablation} isolates the two core components of SkillHEX. Removing self-verification reduces the average pass rate from 55.9\% to 44.8\%, a significant drop of 11.1 pp. The absence of self-supervised dense diagnostic signals severely impairs the reflection module's ability to accurately attribute failures, resulting in performance comparable to SkillRevise \cite{skillrevise}, which similarly lacks self-verification through test cases. Without extracting granular diagnostic signals, the agent struggles to refine its skills effectively. The magnitude of this drop suggests that acquiring discriminative diagnostic evidence is the primary source of improvement in this setting. Furthermore, replacing the skill patch tree with standard in-place refinement (while retaining self-verification) causes a separate 6.8 pp drop, reducing the average pass rate to 49.1\%. This supports the claim that straightforward in-place overwriting tends to trap the agent in local optima without preserving alternative paths for exploration. Nevertheless, this variant still outperforms CoEvoSkill~\cite{coevoskills}, another method that employs a self-verifier, thereby highlighting the intrinsic benefit of our hypothesis-driven approach.

These results align with the intended exploration--exploitation mechanism: hypothesis-driven self-verification ensures highly effective exploitation by squeezing dense diagnostic signals from past interactions, while evidence-guided tree search dynamically enables exploration branches, empowering the agent to escape from  flawed patches and search in promising new directions at appropriate moments. By maximizing the utility of every execution attempt, SkillHEX empowers the agent to sustain skill improvements. 

\subsection{Token Cost}

\begin{table}[t]
\centering
\setlength{\tabcolsep}{3pt}
\begin{tabular}{@{}lrrr@{}}
\toprule
Method & Input & Output & Total \\
\midrule
CoEvoSkills & $2703.7$K & $170.5$K & $2874.3$K \\
SkillRevise & $824.2$K & $38.6$K & $862.7$K \\
\midrule
SkillHEX (full) & $2214.6$K & $141.5$K & $2356.1$K \\
\quad w/o self-verifier & $1100.0$K & $61.3$K & $1161.4$K \\
\quad w/o skill patch tree & $2359.5$K & $142.7$K & $2502.2$K \\
\bottomrule
\end{tabular}
\caption{\textbf{Average per-task token consumption using GPT-5.3-Codex on SkillsBench.} The reported counts include all input and output tokens used for skill evolution and agent execution.}
\label{tab:token_cost}
\end{table}

As shown in Table~\ref{tab:token_cost}, SkillHEX consumes 18.0\% fewer tokens than the strongest baseline, CoEvoSkills, while achieving a substantially higher pass rate. Removing the self-verifier reduces token consumption by roughly half, but decreases performance by 11.1 pp. Notably, replacing the skill patch tree with standard in-place refinement actually increases total token consumption by 6.2\%, while reducing performance by 6.8 pp. This result indicates that in-place refinement may expend additional computation by continuing along an unproductive revision path, whereas tree search can use retrospective test results from the evidence bank to compare revision effects and revisit promising skill versions preserved in the tree, thereby improving both search effectiveness and token efficiency.

\subsection{Effect of Initial Skills}
\begin{figure}[t]
\centering
\includegraphics[width=0.95\columnwidth]{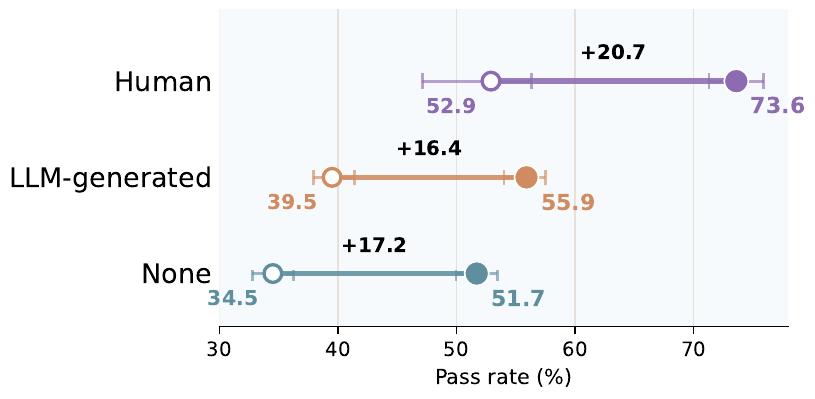}
\caption{\textbf{Effect of Initial Skills.} Performance comparison on SkillsBench using GPT-5.3-Codex after five iterations.}
\label{fig:initialization}
\end{figure}

As shown in Figure~\ref{fig:initialization}, evolution without an initial skill achieves performance close to evolution from an LLM-generated skill, suggesting that SkillHEX can construct useful skills from scratch. Initialization with human-curated skills yields a substantially stronger result. We attribute this stronger result to the fact that human-curated skills provide strong domain-specific and procedural priors that narrow the revision space. Guided by self-verification evidence, the agent can operationalize broad human guidance into task-specific execution specifications, including precise workflows, artifact contracts, numerical conventions, and validation procedures. 
Overall, a strong initialization guides evolution toward more promising revisions and enables higher performance within the same iteration budget.

\section{Conclusion}
We present SkillHEX, a novel closed-loop framework for agent skill evolution under limited budgets and sparse test-time feedback. By coupling a hypothesis-driven self-verifier, which translates ambiguous failures into reusable diagnostic evidence, with a persistent patch tree that allocates evaluations across competing revisions instead of committing to a single edit sequence, SkillHEX effectively prevents premature commitment to flawed edits. Evaluated on 87 SkillsBench tasks, SkillHEX achieves pass rates of 55.9\% and 57.9\%  under a five-iteration budget across two backbone models. This surpasses human-curated skills and outperforms the strongest baseline by 9.5 and 8.5 percentage points, respectively. Ablation and initialization studies confirm that both components provide robust, complementary gains regardless of the starting skill quality. Ultimately, our results demonstrate that actively acquiring discriminative evidence and preserving revision alternatives are essential for efficient test-time adaptation.

\bibliography{aaai2027}

\appendix

\section{Prompt Templates}
\label{app:prompts}
The following boxes reproduce the deployed prompt content in a
paper-readable form. We preserve the role instructions, decision rules, hard
constraints, and output fields. Only repetitive task-family adapters,
machine-specific paths, and examples are condensed.

\begin{promptbox}{System Prompt for Reflection}
\textbf{\# ROLE}

You are the \textbf{attempt-local reflection step} inside one task trial:
run task $\rightarrow$ author public self-verifier tests $\rightarrow$ reflect
$\rightarrow$ optionally gather more evidence $\rightarrow$ emit skill patch
candidates $\rightarrow$ test $\rightarrow$ repeat. In each round, output
either (i) \promptterm{need\_more\_evidence}, together with updates to a dynamic
list of falsifiable hypotheses, or (ii) \promptterm{emit\_patch}, together with
candidate skill patches.

\textbf{\# HARD RULES}

\noindent\textbullet\ Patch operators are only \promptterm{new} and \promptterm{modify}.
\promptterm{new} creates a skill and has empty \promptterm{parent\_skill\_ids};
\promptterm{modify} rewrites exactly one named parent skill.

\noindent\textbullet\ Return \textbf{complete files}: the full \promptterm{SKILL.md} and
every changed script/reference. Never return diffs, ellipses, truncations, or
fragments. Files not returned remain unchanged.

\noindent\textbullet\ Do not repeat an existing hypothesis or an already-tried edit
intent. Official evaluator information is boolean only; never infer or request
additional hidden feedback.

\noindent\textbullet\ Self-verifier tests are advisory probes. They may be incomplete,
overly local, or incorrect; do not treat them as ground truth.

\noindent\textbullet\ If \promptterm{must\_emit\_patch=true}, return
\promptterm{emit\_patch} with at least one candidate even when evidence is weak.

\textbf{\# METHOD}

\textbf{1. Decide sufficiency.}
Judge each existing hypothesis using its attached advisory probes, then choose
one of two decisions.

\noindent\textbullet\ \textbf{Insufficient evidence:} if
\promptterm{must\_emit\_patch=false}, return
\promptterm{need\_more\_evidence}. Emit at least one
\promptterm{hypothesis\_op}, set \promptterm{active\_hypothesis\_ids} to the
hypotheses that the self-verifier should probe next, and emit no patch
candidates.

\noindent\textbullet\ \textbf{Sufficient evidence or mandatory emission:} if
the evidence is sufficient, or \promptterm{must\_emit\_patch=true}, return
\promptterm{emit\_patch} and propose child patch candidates as described in
Step 4.

Keep probing hypotheses that remain plausible under official failure. Refine
hypotheses whose probes expose only a local symptom, and refute only those that
the probes genuinely invalidate or render non-discriminating. When requesting
more evidence, state both the hypothesis and the observable behavior that
would support or refute it.

\textbf{2. Maintain the hypothesis list.}
Keep the list diverse so distinct failure directions do not collapse into one
guess.

\noindent\textbullet\ \promptterm{add}: introduce a falsifiable claim on a
distinct failure surface.

\noindent\textbullet\ \promptterm{refine} (\promptterm{hypothesis\_id}): narrow
or rewrite an existing claim in place.

\noindent\textbullet\ \promptterm{refute} (\promptterm{hypothesis\_id}): mark
a wrong, hallucinated, unsupported, or non-discriminating claim as refuted.
Prune its test columns only when no active hypothesis also references them.

\textbf{3. Route hypotheses.}

\textbf{Hard deliverable contract (probe first).} Check exact output paths;
artifact existence and parseability; required keys, columns, sheets, pages,
slides, rows, records, or counts; field types; API signatures; output wrappers;
and formula requirements. Formulate the hypothesis as: ``The answer violates
an explicit deliverable contract from the task instruction.'' Ask for tests
that compare the answer with the public instruction, never hidden
expectations.

\textbf{Content or process quality.} If existing cases already cover the
relevant contract and runtime surfaces and pass, do not continue generating
the same contract hypotheses. Instead, probe whether the produced artifacts
correctly solve the task, including feasibility, numerical conventions,
coverage, objective quality, preservation, runtime behavior, or rendering.

\textbf{4. Propose patch candidates.}
Emit candidates as children of the current node; search allocates rollouts
using their rank-derived prior and observed reward.

\noindent\textbullet\ \textbf{One root cause per candidate.} Target one
coherent local, mechanism-level, or systemic cause. Include every change that
the selected cause requires, but do not bundle unrelated fixes.

\noindent\textbullet\ \textbf{Ordinal rank.} List candidates best-first, with
rank 1 denoting the most promising candidate.

\noindent\textbullet\ \textbf{Wholesale modification.} A
\promptterm{modify} candidate must re-derive the complete instruction or
algorithm from public evidence and remove assumptions, formulas, thresholds,
or conventions that the evidence refutes; it must not merely append text.

\noindent\textbullet\ \textbf{Alternative branches.} Propose multiple
self-contained and independently testable candidates when several distinct
root causes remain plausible.

\textbf{\# OUTPUT CONTRACT}

Return exactly one JSON object with the following structure:
\begin{contractbox}{Reflection JSON Schema}
\contractfield{decision}{One of \promptterm{need\_more\_evidence} or
\promptterm{emit\_patch}.}
\contractfield{sufficiency}{Object containing \promptterm{is\_sufficient},
\promptterm{confidence}, and \promptterm{reason}.}
\contractfield{hypothesis\_ops}{List of operations. Each item contains
\promptterm{op} (\promptterm{add}, \promptterm{refine}, or
\promptterm{refute}), \promptterm{hypothesis\_id}, \promptterm{text},
\promptterm{target\_behavior}, and \promptterm{reason}.}
\contractfield{active\_hypothesis\_ids}{IDs selected for the next
self-verification round.}
\contractfield{patch\_candidates}{Ranked list of candidates. Each item contains
\promptterm{patch\_operator}, \promptterm{hypothesis},
\promptterm{parent\_skill\_ids}, \promptterm{evidence\_case\_ids},
\promptterm{rank}, \promptterm{edit\_intent}, \promptterm{skill\_patch}, and
\promptterm{notes}.}
\contractfield{edit\_intent}{Object containing
\promptterm{primary\_failure\_mode}, \promptterm{target\_behavior},
\promptterm{non\_goals}, and \promptterm{rationale}.}
\contractfield{skill\_patch}{Complete package containing \promptterm{name},
\promptterm{description}, \promptterm{skill\_md}, \promptterm{scripts}, and
\promptterm{references}. Each auxiliary file provides its path and full
content.}
\contractfield{summary}{Short summary of the reflection round.}
\end{contractbox}
\end{promptbox}

\begin{promptbox}{System Prompt for Self-Verifier}
\textbf{\# ROLE}

You are the \textbf{SELF-VERIFIER} for an autonomous coding agent, working
inside a Linux container. Given one or more reflection hypotheses, author
public executable tests that distinguish whether each hypothesis is true for
the restored agent answer, then leave the tests on disk.

\textbf{\# HARD RULES}

\textbullet\ Use public information only: the task instruction, visible task
inputs, restored answer artifacts, their manifest, and public runtime tools.
Never use hidden tests or guessed oracles.

\textbullet\ Expected values must be independently derived from the instruction,
public inputs, answer artifacts, canonical tools, or runtime probes. If a check
uses a heuristic, subjective threshold, or proxy for hidden quality, label it
\promptterm{proxy\_quality} or \promptterm{diagnostic}.

\textbullet\ Each test checks one property, is deterministic and recomputable,
and reports exactly \promptterm{SELF\_VERIFIER\_RESULT=PASS} or
\promptterm{SELF\_VERIFIER\_RESULT=FAIL}. A test may be dry-run only to confirm
that it executes; the harness records the final rerun.

\textbullet\ Never print an entire large data or media file. Inspect only
schemas, shapes, counts, headers, metadata, aggregates, samples, or tool
summaries, and place any full scan inside the authored test script.

\textbf{\# WORKFLOW}

1. Discover the runtime layout and locate public inputs and restored outputs.
Use the answer manifest rather than assuming optional paths.

2. Inspect reductions of the real artifacts needed by the hypothesis. Do not
explore the entire task or generate unrelated task-family checks.

3. For each active hypothesis, author one or two high-signal cases using the
strongest available public oracle. Set \promptterm{hypothesis\_ids} to the exact
hypothesis or hypotheses probed. A useful failure message must identify the
root cause and a concrete repair direction.

4. Write each draft below the supplied
\promptterm{current\_case\_root/<test\_id>/}. When validator feedback is supplied,
repair or replace invalid cases and leave already valid cases unchanged. Stop
after creating the requested number of cases and summarize their IDs.

\textbf{\# ASSERTION STRENGTH}

Use exactly one of the following labels:

\textbullet\ \promptterm{hard\_contract}: an explicit path, schema, format,
cardinality, signature, preservation, formula, or artifact requirement.

\textbullet\ \promptterm{environment\_preflight}: import, build, typecheck,
server startup, command availability, or declared runtime compatibility.

\textbullet\ \promptterm{deterministic\_oracle}: a result independently recomputed
from public inputs or a canonical public tool.

\textbullet\ \promptterm{proxy\_quality}: a public baseline, bound, threshold,
robustness, performance, or quality proxy that is not hidden truth.

\textbullet\ \promptterm{diagnostic}: an ambiguity probe, heuristic suspicion,
convention comparison, or other weak signal used only for reflection.

\textbf{\# CASE OUTPUT CONTRACT}

For every test, create
\promptterm{<current\_case\_root>/<test\_id>/case.json} and the executable
script named by \promptterm{command}. The JSON object has the following fields:

\begin{contractbox}{Self-Verifier Case Schema}
\contractfield{test\_id}{Unique identifier of the form
\promptterm{t\_<short\_slug>}.}
\contractfield{test\_type}{One of \promptterm{file\_exists},
\promptterm{schema}, \promptterm{format}, \promptterm{small\_example},
\promptterm{invariant}, \promptterm{consistency}, \promptterm{runtime},
\promptterm{quality}, or \promptterm{other}.}
\contractfield{assertion\_strength}{One of \promptterm{hard\_contract},
\promptterm{environment\_preflight}, \promptterm{deterministic\_oracle},
\promptterm{proxy\_quality}, or \promptterm{diagnostic}.}
\contractfield{oracle\_source}{Short label identifying the public source of
the expected value.}
\contractfield{public\_support}{Concrete public evidence supporting the test.}
\contractfield{purpose / summary}{Human-readable purpose and a self-contained
description of what is checked and how.}
\contractfield{answer.expected}{Expected behavior asserted by the test.}
\contractfield{diagnosis}{Object containing \promptterm{failure\_mode} and
\promptterm{repair\_hint}.}
\contractfield{command}{Standalone execution command, e.g.,
\promptterm{python3 test.py}.}
\contractfield{hypothesis\_ids}{Exact IDs of the hypotheses probed by this
test.}
\end{contractbox}

The fields \promptterm{public\_support}, \promptterm{summary},
\promptterm{failure\_mode}, and \promptterm{repair\_hint} must be concrete enough for
reflection to understand the evidence and repair implication without reading
the test source. For multiple hypotheses, create at least one dedicated test
per hypothesis whenever public evidence permits it.
\end{promptbox}

\section{Experimental Details}
\label{app:proofs}

\subsection{Experimental Configuration}
\label{app:algorithm-configuration}
Table~\ref{tab:skillhex-defaults} reports the fixed configuration used for all SkillHEX main experiments. The attempt budget allows at most $K=5$ evolved skill executions following the initial root evaluation (maximum six total executor calls). Node expansion retains up to $B=5$ children. Candidate ranks are converted to priors using temperature $\beta=1.0$, and tree traversal uses $c_{\rm puct}=1.4$. Unevaluated children receive a first-play-urgency (FPU) equal to their parent's normalized value minus $0.1$.

\begin{table}[t]
\centering
\caption{Default SkillHEX configuration. The new-case cap applies per self-verifier invocation, not to the overall persistent test bank size.}
\label{tab:skillhex-defaults}
\small
\begin{tabular}{@{}p{0.20\columnwidth}p{0.43\columnwidth}p{0.25\columnwidth}@{}}
\toprule
Component & Parameter & Value \\
\midrule
Executor & Evolved-attempt budget $K$ & 5 \\
Patch expansion & Maximum children $B$ & 5 \\
Evidence loop & Reflection rounds $L$ & 3 \\
Self-verifier & New-case cap $C$ & 10 \\
Self-verifier & Validation/repair rounds $V$ & 3 \\
Search & Rank-prior temperature $\beta$ & 1.0 \\
Search & PUCT constant $c_{\rm puct}$ & 1.4 \\
Search & First-play-urgency reduction & 0.1 \\
Agents & Executor / verifier step caps & 10,000 / 1,000 \\
Test execution & Per-case timeout & 180 seconds \\
Case audit & LLM verification of new cases & Enabled \\
Patch space & Allowed operators & \texttt{new}, \texttt{modify} \\
\bottomrule
\end{tabular}
\end{table}

The evidence loop alternates reflection and self-verification for up to $L=3$ rounds per node expansion. The loop terminates immediately once reflection emits a valid patch. Otherwise, hypothesis operations (Add/Refine/Refute) update the hypothesis store, and the self-verifier probes active hypotheses. In the final round, \texttt{must\_emit\_patch=true} is enforced; failing to emit a valid patch marks the node as exhausted, redirecting the search to another frontier.

Each self-verifier invocation admits up to $C=10$ new valid cases, utilizing one generation round and up to $V-1=2$ repair rounds based on validator feedback. Cases remaining invalid after round $V$ are discarded. Accepted cases are executed (180-second timeout), audited, and replayed across cached node outputs to update the shared evidence matrix. Validated tests from prior rounds remain available in a persistent bank for future replay. All parameters in Table~\ref{tab:skillhex-defaults} are fixed across tasks and models, except executor timeouts, which follow benchmark-specific limits.

\subsection{Executor Harness}

To isolate the impact of evolved skills from framework-level interventions, we employ a deliberately lightweight executor harness. Its design enforces strict isolation and evaluation integrity through three core mechanisms:

\paragraph{Strict Environment Isolation.} For every execution attempt, the harness destroys the previous container and launches a fresh instance from the base task image, preventing any cross-attempt leakage of files, dependencies, or process states. It injects only the selected skill revision. Consequently, the staged skill package remains the sole experimental variable, while the base model, task instruction, and compute limits are held strictly constant.

\paragraph{Minimal Action Interface.} The executor agent operates under a basic system prompt instructing it to utilize the installed skills. It is equipped with only two generic tools: \texttt{run\_bash} (for executing shell commands and retrieving truncated standard streams and exit statuses) and \texttt{finish}. The agent has no access to task-specific APIs, intermediate reward shaping, reflection feedback, self-verifier verdicts, or hidden tests during execution.

\paragraph{Rigorous Evaluation Boundary.} Upon termination, the harness exports all generated artifacts from designated output directories before recycling the container. The benchmark's official verifier evaluates these artifacts and returns only a sparse binary success signal to the evolution policy. Detailed verifier outputs are strictly hidden from all LLMs (executor, reflection, and self-verifier prompts) to preserve the hidden evaluation boundary. 

\section{Evolution on Human-curated Skills}

\begin{figure}[t]
\centering
\includegraphics[width=1.0\columnwidth]{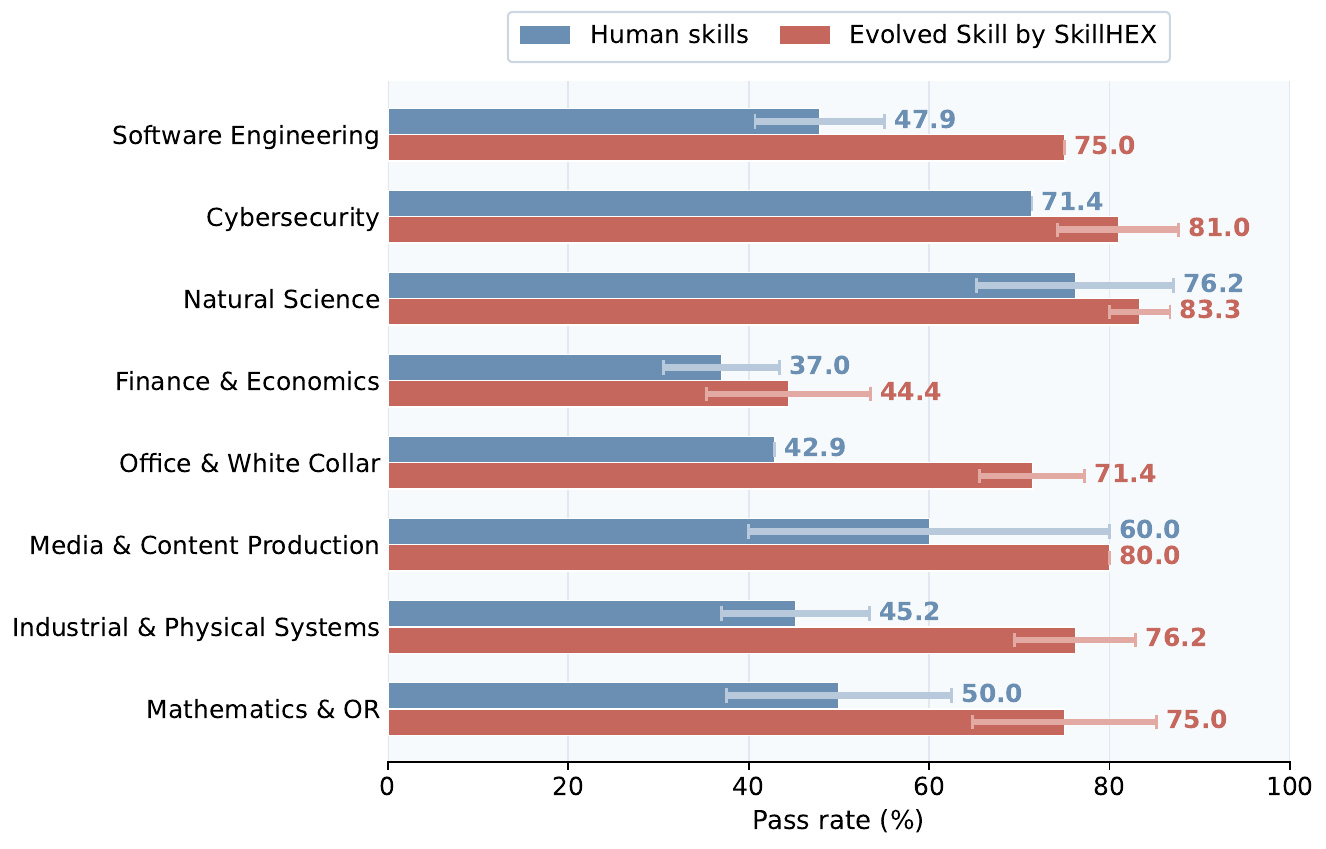}
\caption{Domain-level effect of evolving human-curated skills with SkillHEX on SkillsBench using GPT-5.3-Codex SkillHEX improves the mean pass rate in domains under 5 iterations.}
\label{fig:human-skill-domain-comparison}
\end{figure}

To complement the initialization analysis in the main paper, we examine how SkillHEX evolves the human-curated skills within each domain. Figure~\ref{fig:human-skill-domain-comparison} compares the original human-skill initialization with the resulting SkillHEX performance; bars report the mean pass rate over three seeds and error bars show the
corresponding standard deviation.

The domain-wise evolution of human-curated skills reveals a symbiotic relationship between human strategic priors and autonomous test-time adaptation. While human-curated skills provide robust procedural baselines and effectively narrow the initial search space, static guidelines inherently leave a refinement gap regarding low-level, task-specific constraints such as implicit numerical conventions and strict artifact contracts. SkillHEX bridges this gap by operationalizing broad human guidance into rigorous, environment-grounded execution specifications via hypothesis-driven self-verification. This mechanism explains the substantial performance leaps in workflow-heavy domains like Software Engineering and Operations Research, where procedural precision is critical. Conversely, in knowledge-intensive fields like Natural Science and Finance, evolutionary gains remain bounded. As noted in the main paper, failures in these domains stem primarily from a knowledge-acquisition bottleneck, missing specialized facts that cannot be inferred purely from execution traces. Ultimately, by leveraging human-curated skills as strong procedural priors, SkillHEX effectively narrows the exploration space for skill modifications, enabling the search mechanism to efficiently discover high-performing, task-specific revisions.

\section{Case Study}
\begin{figure}[t]
\centering
\includegraphics[width=1.0\columnwidth]{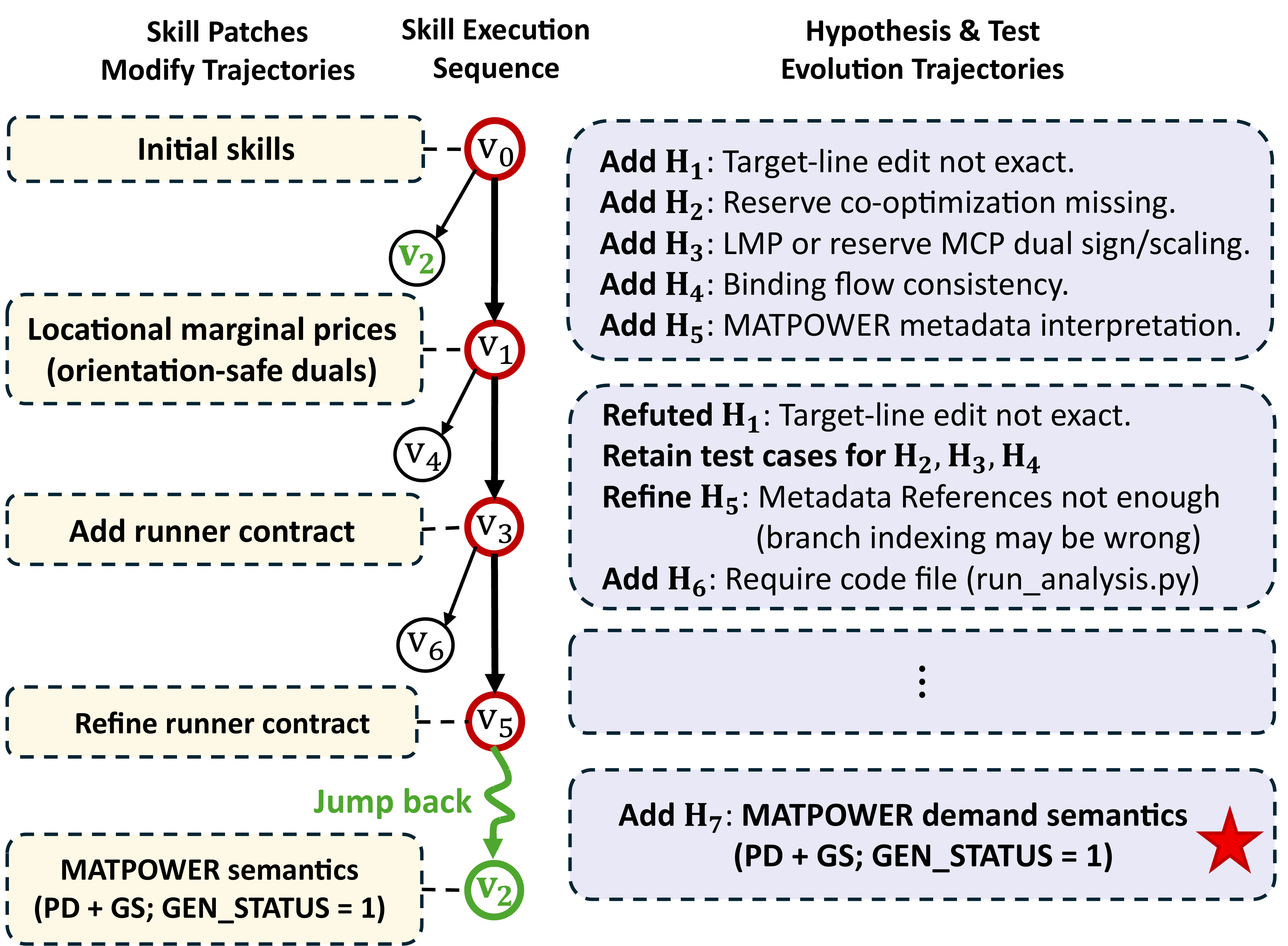} 
\caption{SkillHEX trajectory on \textbf{energy-market-pricing} task. Red, green, and gray node outlines represent failed, successful, and unevaluated retained skill candidates, respectively. The left yellow boxes summarize the modifications evaluated at each node, while the right purple boxes display the corresponding hypothesis operations. The faded copy of $V_2$ next to the root denotes the lower-priority candidate preserved when $V_1$ was initially selected. The red dashed arrow indicates a tree-search backtracking jump from the unsuccessful branch ($V_1\!\rightarrow\!V_3\!\rightarrow\!V_5$) back to this candidate based on reward updates. This jump discards the inherited LMP and runner-contract edits, ultimately yielding the successful MATPOWER-semantics revision.}
\label{fig:energy-market-case}
\end{figure}

We use the \textbf{energy-market-pricing} task to illustrate SkillHEX's ability to navigate complex failure landscapes. The agent must solve a MATPOWER-based DC optimal power flow problem with reserve co-optimization, reporting locational marginal prices (LMPs) and a reserve market-clearing price (MCP). This creates an ambiguous credit assignment problem: failures may stem from output contracts, dual scaling errors, or underlying MATPOWER data semantics. Figure~\ref{fig:energy-market-case} traces a five-attempt trajectory where the search escapes a local optimum by backtracking to an alternative hypothesis preserved at the root.

At the root ($V_0$), the initial answer passed six basic public checks but yielded an uninformative official failure. Acknowledging this ambiguity, SkillHEX formulated five distinct hypotheses ($H_1$--$H_5$) and initially prioritized $V_1$ to address potential LMP dual scaling errors. To refine its diagnosis without consuming execution budgets, SkillHEX leveraged reusable tests on cached outputs. This pruned the hypothesis space and introduced a new runner-contract hypothesis ($H_6$), driving the search down the $V_1\!\rightarrow\!V_3\!\rightarrow\!V_5$ branch. While internal self-verifier scores climbed (from 0.3462 to 0.4231) as contract probes passed, core semantic tests continued to fail, and the official verifier rejected the final node. The dense internal evidence had inadvertently created a local optimum trap.

A traditional greedy in-place method would fatally stall here, dragging the ineffective LMP and contract edits into all subsequent attempts. SkillHEX, however, executed an evidence-guided branch switch to $V_2$, the lower-ranked sibling preserved at the root. This transition provided a clean slate, discarding the inherited baggage of the previous branch. The revision at $V_2$ focused exclusively on making MATPOWER data-model semantics explicit ($H_7$): enforcing strict \texttt{GEN\_STATUS=1} filtering and properly coupling energy-reserve capacities. This targeted semantic alignment passed the official verifier on the final attempt. Ultimately, this demonstrates how hypothesis-driven self-verification sharpens \textit{what} to try, while the persistent patch tree guarantees \textit{where} to try it when the most promising path fails.

\section{Algorithm}
\begin{algorithm}[H]
\caption{SkillHEX}
\label{alg:skillhex}
\begin{algorithmic}[1]
\Require Task $\mathcal{T}$ and initial skill $S^{(0)}$, Evolved-attempt budget $K$, branch width $B$, evidence-round limit $L$, Case cap $C$, validation-round limit $V$, prior temperature $\beta$, constant $c_{\rm puct}$
\Ensure best evolved skill $S^{\ast}$
\State Initialize patch tree $\mathcal{G}\leftarrow\{v_0:S^{(0)}\}$
\State Initialize hypotheses $\mathcal{H}$, tests $\mathcal{C}$, and evidence $\mathbf{M}$ as empty
\State $(Y_{v_0},R_{v_0})\leftarrow\Call{Execute}{\mathcal{T},S^{(0)}}$
\If{$R_{v_0}=1$} \Return $S^{(0)}$ \EndIf
\State $(\Omega,\mathcal{A},\mathcal{P})\leftarrow\Call{Reflect}{v_0,\mathcal{H},\mathbf{M},\mathrm{false}}$ \Comment{\textcolor{gray}{initial hypotheses}}
\State $\mathcal{H}\leftarrow\Call{UpdateHypotheses}{\mathcal{H},\Omega}$ 
\State $\Call{SelfVerify}{v_0,\mathcal{A},C,V}$ \Comment{\textcolor{gray}{updates $\mathcal{C},\mathbf{M}$}}
\State $(s,Q)\leftarrow\Call{EvaluateAndBackup}{\mathcal{G},\mathbf{M}}$
\State $k\leftarrow 0$
\While{$k<K$}
  \State $v\leftarrow\Call{Select}{\mathcal{G},Q,P,N,c_{\rm puct}}$ \Comment{\textcolor{gray}{Eq.~(10)}}
  \If{$v=\mathrm{none}$} \textbf{break} \EndIf
  \If{$v$ is unevaluated}
    \State $(Y_v,R_v)\leftarrow\Call{Execute}{\mathcal{T},S_v}$; $k\leftarrow k+1$
    \If{$R_v=1$} \Return $S_v$ \EndIf
    \State $\Call{ReplayTests}{v,\mathcal{C}}$ \Comment{\textcolor{gray}{evaluate against history; Eq.~(5)}}
  \Else \Comment{\textcolor{gray}{hypothesis-driven expansion}}
    \State $\mathcal{P}\leftarrow\varnothing$
    \For{$\ell=1$ \textbf{to} $L$}
      \State $m\leftarrow[\ell=L]$ \Comment{\textcolor{gray}{force emission in the final round}}
      \State $(\Omega,\mathcal{A},\mathcal{P})\leftarrow\Call{Reflect}{v,\mathcal{H},\mathbf{M},m}$
      \State $\mathcal{H}\leftarrow\Call{UpdateHypotheses}{\mathcal{H},\Omega}$ \Comment{\textcolor{gray}{Add/Refine/Refute}}
      \State $\Call{PruneRefuted}{\mathcal{H}}$ \Comment{\textcolor{gray}{updates $\mathcal{C},\mathbf{M}$; Eq.~(7)}}
      \If{$\mathcal{P}\neq\varnothing$} \textbf{break} \EndIf \Comment{\textcolor{gray}{break after hypothesis updates}}
      \State $\Call{SelfVerify}{v,\mathcal{A},C,V}$ \Comment{\textcolor{gray}{generate new tests; Eqs.~(5)--(6)}}
    \EndFor
    \If{$\mathcal{P}=\varnothing$} mark $v$ exhausted; \textbf{continue} \EndIf
    \State $\mathcal{P}_B\leftarrow\Call{TopB}{\mathcal{P}}$
    \State $\mathcal{G}\leftarrow\Call{Expand}{\mathcal{G},v,\mathcal{P}_B}$
    \State $P(v,\cdot)\leftarrow\Call{RankPrior}{\mathcal{P},\beta}$ \Comment{\textcolor{gray}{Eq.~(8)}}
  \EndIf
  \State $(s,Q)\leftarrow\Call{EvaluateAndBackup}{\mathcal{G},\mathbf{M}}$ \Comment{\textcolor{gray}{Eq.~(9)}}
\EndWhile
\State $v^\ast\leftarrow\Call{SelectBest}{\mathcal{G},s}$ \Comment{\textcolor{gray}{among evaluated nodes}}
\State \Return $S_{v^\ast}$
\end{algorithmic}
\end{algorithm}

\end{document}